\pdfoutput=1

\documentclass[11pt]{article}

\PassOptionsToPackage{dvipsnames,usenames}{xcolor}
\usepackage[final]{acl}

\usepackage{times}
\usepackage{latexsym}

\usepackage[T1]{fontenc}

\usepackage[utf8]{inputenc}

\usepackage{microtype}

\usepackage{graphicx}
\usepackage{dsfont}
\usepackage{caption}
\usepackage{multirow}
\usepackage{subcaption}
\usepackage{soul}
\usepackage{hyperref}
\usepackage{adjustbox}
\usepackage{booktabs}
\usepackage{makecell}
\usepackage{nicefrac}
\usepackage{xspace}
\usepackage{amsfonts}
\usepackage{amsthm}
\usepackage{amsmath}
\usepackage{amssymb}
\usepackage{amsfonts}
\usepackage{mathtools}
\usepackage{bm}
\usepackage{bbm}
\usepackage{mleftright}
\usepackage{listings}
\usepackage{rotating}
\usepackage{hyperref}
\usepackage{wrapfig}
\usepackage{geometry}
\DeclareMathOperator*{\E}{\mathbb{E}}

\usepackage[textsize=tiny]{todonotes}

\newcommand{\eg}{\textit{e.g.}}
\newcommand{\ie}{\textit{i.e.}}

\usepackage[capitalize,noabbrev]{cleveref}
\crefname{section}{\S}{\S\S}
\crefname{table}{Tab.}{Tab.}
\crefname{figure}{Fig.}{Figs.}
\crefname{algorithm}{Alg.}{}
\crefname{equation}{Eq.}{Eq.}
\crefname{appendix}{App.}{}
\crefname{theorem}{Theorem}{}
\crefname{prop}{Proposition}{}
\crefname{definition}{Def.}{}
\crefname{cor}{Corollary}{}
\crefname{observation}{Observation}{}
\crefname{assumption}{Assumption}{}
\crefname{hyp}{Hyp.}{Hypotheses}
\crefformat{section}{\S#2#1#3}
\crefname{namedtheorem}{Hyp.}{Hypotheses}
\crefformat{footnote}{#2\footnotemark[#1]#3}

\newcommand{\mymacro}[1]{#1} %

\usepackage{colortbl}
\definecolor{ETHBlue}{RGB}{33,92,175}   %
\definecolor{ETHGreen}{RGB}{98,115,19}      %
\definecolor{ETHPurpleDark}{RGB}{140,10,89} %
\definecolor{ETHPurple}{RGB}{163,7,116} %
\definecolor{ETHGray}{RGB}{111,111,111} %
\definecolor{ETHRed}{RGB}{183,53,45}    %
\definecolor{ETHPetrol}{RGB}{0,120,148} %
\definecolor{ETHBronze}{RGB}{142,103,19}    %
\definecolor{ETHOrange}{RGB}{230, 100, 50}

\colorlet{MacroColor}{ETHGreen}
\colorlet{MACROCOLOR}{MacroColor}

\DeclareMathOperator*{\argmax}{{argmax}}

\newcommand{\indicator}[1]{{\mymacro{\mathbbm{1}\{#1\}}}}

\newcommand{\alphabet}{{\mymacro{\Sigma}}}

\newcommand{\kleene}[1]{#1^*}

\newcommand{\plm}{\mymacro{P}} %

\newcommand{\eos}{\mymacro{\textsc{eos}}}

\newcommand{\dataset}{\mymacro{\mathcal{D}}}
\newcommand{\embed}{\mymacro{\phi}}
\newcommand{\utility}{\mymacro{u}}
\newcommand{\utilitycut}{\mymacro{\utility_{\text{cut}}}}
\newcommand{\cutoffconstant}{\mymacro{\delta}}
\newcommand{\threshold}{\mymacro{\tau}}
\newcommand{\candidates}{\mymacro{\mathcal{H}}}
\newcommand{\cluster}{\mymacro{C}}

\newcommand{\tablefoursize}{\fontsize{7.75pt}{9.1pt}\selectfont}

\title{Structure-Conditional Minimum Bayes Risk Decoding}

\author{Bryan Eikema\textsuperscript{$\diamond$}, Anna Rutkiewicz\textsuperscript{$\star$}, Mario Giulianelli\textsuperscript{$\ltimes$} 
\\
         \textsuperscript{$\diamond$}University of Amsterdam, \textsuperscript{$\star$}University of Zurich, \textsuperscript{$\ltimes$}UCL \\
         \texttt{\href{mailto:b.eikema@uva.nl}{b.eikema@uva.nl} \href{mailto:annamonika.rutkiewicz@uzh.ch}{annamonika.rutkiewicz@uzh.ch}}\\
         \texttt{\href{mailto:m.giulianelli@ucl.ac.uk}{m.giulianelli@ucl.ac.uk}}
         }

\begin{document}
\maketitle

\begin{abstract}

Minimum Bayes Risk (MBR) decoding has seen renewed interest as an alternative to traditional generation strategies.
While MBR has proven effective in machine translation, where the variability of a language model's outcome space is naturally constrained, it may face challenges in more open-ended tasks such as dialogue or instruction-following. 
We hypothesise that in such settings, applying MBR with standard similarity-based utility functions may result in selecting responses that are broadly representative of the model's distribution, yet sub-optimal with respect to any particular grouping of generations that share an underlying \textit{latent structure}. 
In this work, we introduce three lightweight adaptations to the utility function, designed to make MBR more sensitive to structural variability in the outcome space. 
To test our hypothesis, we curate a dataset capturing three representative types of latent structure: dialogue act, emotion, and response structure (\eg, a sentence, a paragraph, or a list). We further propose two metrics to evaluate the structural optimality of MBR. 
Our analysis demonstrates that common similarity-based utility functions fall short by these metrics. In contrast, our proposed adaptations considerably improve structural optimality. 
Finally, we evaluate our approaches on real-world instruction-following benchmarks, AlpacaEval and MT-Bench, and show that increased structural sensitivity improves generation quality by up to 13.7 %
percentage points in win rate.\looseness-1\footnote{We release experimental code and the structural variation dataset at \url{https://github.com/Roxot/structure-conditional-mbr}.}

\end{abstract}

\section{Introduction}
Once a language model has been trained, one fundamental problem remains: determining how to select an output sequence from the model's learned probability distribution over possible continuations, given a particular context. 
Traditional approaches such as beam search decoding and majority voting aim to select a high probability continuation under the model distribution. %
However, a growing body of research has shown that model probability does not reliably align with human preferences \citep{stahlberg-byrne-2019-nmt,zhang-etal-2021-trading} and, in response, Minimum Bayes Risk \citep[MBR;][]{kumar-byrne-2004-minimum,eikema-aziz-2020-map} decoding has emerged as a more robust alternative. %
MBR casts decoding as a decision-theoretic problem, where the selected sequence is the one that minimises risk with respect to a task-specific utility function, under the uncertainty over continuations represented by the language model. This utility typically reflects the degree of agreement between a candidate and the broader set of outcomes, %
penalising candidates that diverge significantly from the consensus. By integrating both model probabilities and inter-candidate consistency, MBR yields generations that are better aligned with human preferences, regularly outperforming conventional methods \citep{freitag-etal-2022-high,prometheus-mbr}. %

\begin{figure}[t]
    \centering
    \includegraphics[width=0.48\textwidth]{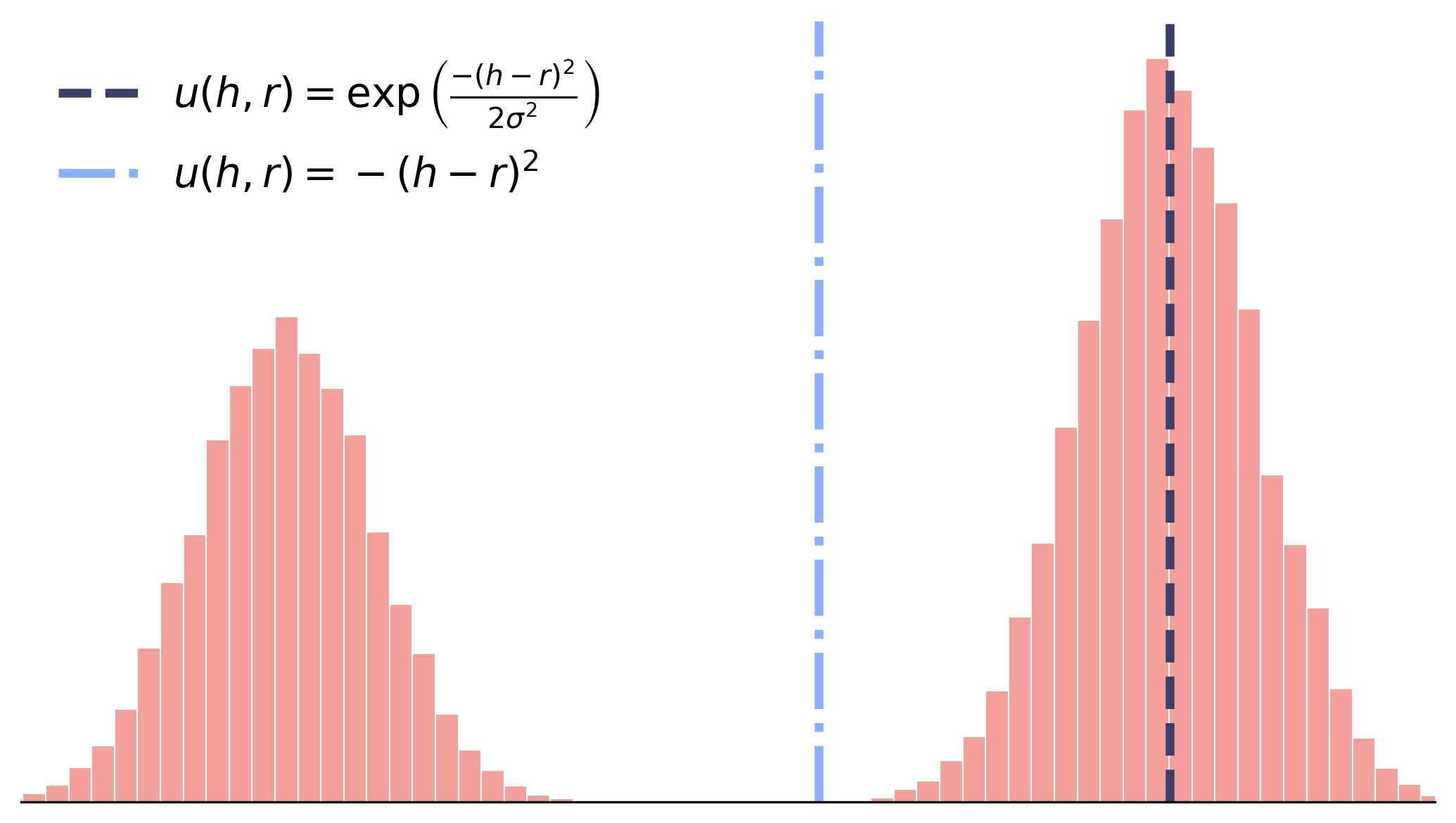}
        \caption{The choice of utility function can considerably impact the Minimum Bayes Risk optimum. When the outcome space is structured or multimodal, the MBR optimum may settle between modes, landing in a region of low probability. Here, we present a continuous example featuring a bimodal Gaussian distribution and show the MBR optima (dashed vertical lines) of two utility functions with markedly different behaviours.}
    \label{fig:continuous-bimodal-mbr}
    \vspace{-0.8em}
\end{figure}

MBR decoding has gained significant attention in neural machine translation, where utility is often measured by task-agnostic sentence similarity scores. %
This corresponds to selecting the sequence which, in expectation and under the lens of a particular similarity score, most closely matches the broader distribution of sequences prescribed by the model. %
While this decoding strategy works well for translation, where outcome space variability is inherently constrained by the task, it
risks being less effective for tasks with a broader range of contextually plausible \textit{latent structures}---and thus greater variability in realisations---such as dialogue or instruction-following \cite{giulianelli-etal-2023-comes}. %
Consider, for instance, the following dialogue exchange: \textbf{A:} \textit{The mountains would be a great place for the lab retreat.} \textbf{B:} \textit{That's a wonderful choice.} In response, speaker~A could follow up with a statement (\textit{The mountains offer many outdoor team-building activities.}), a question (\textit{Which aspects of the mountains are you most excited about?}), a directive (\textit{Please check out different venues online to finalise the decision.}), or an offer (\textit{Shall I make the necessary arrangements?}). Similarly, when given an instruction like \textit{Please summarise Gödel, Escher, Bach}, valid responses could range from a single-sentence summary to a detailed paragraph, a multi-paragraph narrative, or even a list of key topics.
In such settings, applying MBR with a standard similarity-based utility function may result in selecting an output that is broadly representative of the model's outcome distribution, but suboptimal with respect to any one plausible latent structure (we illustrate this in \cref{fig:continuous-bimodal-mbr} using a continuous distribution as a simplified example, and in \cref{tab:motivating-example-generations} using a real-world example with a customer service language model).  %

In this work, we propose %
adapting utility functions for MBR such that they are able to explicitly account for a language model's uncertainty over latent structures. %
We adopt a broad interpretation of \emph{structure}, treating it as a latent variable that influences the form a generation takes, such as a dialogue act, the level of detail in a response, %
or the emotion conveyed by an utterance. 
To examine how reliably MBR selects the highest-consistency candidate within clusters of generations that share a latent structure---what we call \textit{cluster-optimality}---we semi-automatically construct a dataset of 3,000 curated outcome spaces, %
for a total of 350,000 candidate generations.
These are conditioned on naturally occurring conversational and instruction-following contexts, but present controlled uncertainty over three types of structure: dialogue act, emotion, and response structure (\ie, a single sentence, a paragraph, a list, or a table).
Our analysis of this dataset shows that, under commonly used utility functions, MBR solutions are cluster-optimal \textit{in fewer than half of the cases}. %
To address this, we introduce three new approaches---Clustering, Structure Embeddings, and Utility Cut-off---that adapt utility functions to account for a candidate's (soft) membership in structure-specific candidate groups, while preserving the decision-theoretic foundation of risk minimisation. %

Our experiments confirm that adapting the utility function to account for latent structural variability substantially improves MBR solutions. 
On our curated dataset with controlled uncertainty over dialogue act, emotion, and response structure, our three proposed methods achieve markedly higher cluster optimality than standard MBR with BERTScore or BLEURT utilities.
We also observe gains on real-world instruction-following benchmarks, demonstrating that our methods can uncover and exploit latent structural variability even without explicit structure annotations.
In particular, our methods improve generation quality on AlpacaEval and MT-Bench, with win rates against %
GPT-4o increasing by up to 13.7 %
percentage points on the latter.
These findings support our central claim: structure-aware utility functions enable MBR to more reliably select high-quality sequences in tasks where structural variability is inherent to the outcome space.

\section{Language Modeling and Decision Rules} %

A language model $\plm$ is a distribution over strings $\kleene{\alphabet}$, where $\alphabet$ is an alphabet, \ie, a finite, non-empty set of symbols, and  $\kleene{\alphabet}$ its Kleene closure, \ie, the set of all strings formed by concatenating symbols in $\alphabet$, including the empty string $\varepsilon$. We define $Y$ as a random variable over sequences in $\kleene{\alphabet}$.
Every language model can be expressed in autoregressive form by decomposing the probability of a string $y \in \kleene{\alphabet}$ as the product of conditional probabilities of each of its symbols, followed by an end-of-string event $\eos$: %
\begin{equation}
\label{eq:autoregressive}
P(Y=y) = P(\eos \mid y) \prod_{t = 1}^{|y|} P(y_t|y_{<t})
\end{equation}
where each conditional distribution $P\mleft(Y_t \mid y_{<t}\mright)$ is a probability distribution over $\alphabet \cup \{\eos\}$.
This formulation underlies most modern autoregressive language models, where each conditional probability is produced by a learned parametric model. We assume an implicit conditioning on a set of neural network parameters $\theta$, estimated during training on a given dataset. Furthermore, because language models are commonly conditioned on an input, or a prompt, $x \in \kleene{\alphabet}$, we are typically only interested in the conditional probability distribution over responses $P(Y|x)$. In the rest of this work, we will always assume the presence of such an input $x$, such as an instruction or dialogue history.

\subsection{Decision Rules}
To obtain a generation from a trained language model $P$, given some input $x$, it is necessary to decide on a single ``best'' outcome in $\kleene{\alphabet}$. Formally, this requires a decision rule that defines a mapping from a distribution $P$ to such an outcome~$y^*$. A common choice is to output the highest probability outcome under $P(Y|x)$, a decision rule known as \textit{maximum-a-posteriori}, typically approximated using beam search or majority voting: %
\begin{equation}
    y^*_\textrm{MAP} = \argmax_{h\in\kleene{\alphabet}} P(Y=h|x)
\end{equation}
However, studies have shown that model probability does not reliably align with human preferences \citep{stahlberg-byrne-2019-nmt,zhang-etal-2021-trading}, and Minimum Bayes Risk (MBR) has become a popular alternative. 
MBR stems from the principle of maximisation of expected utility \citep{BergerSDT}. It requires choosing a \textit{utility} function $\utility(h, r)$ that measures the benefit of choosing hypothesis $h$ given an ideal decision $r$. In natural language generation, $\utility$ is typically chosen to be a strong sentence similarity metric such as BLEURT \citep{sellam-etal-2020-bleurt,freitag-etal-2022-high}, COMET \citep{rei-etal-2020-comet,fernandes-etal-2022-quality} or BERTScore \citep{bertscore,suzgun-etal-2023-follow}. MBR then selects the outcome maximising utility in expectation under the model distribution:
\begin{equation}
    y^*_\textrm{MBR} = \argmax_{h\in\kleene{\alphabet}} \E_{P(Y|x)}\left[\utility(h, Y)\right]
\end{equation}
A sampling-based approximation of MBR has recently gained popularity. It generates a set of unbiased samples from the model and ranks them using Monte Carlo estimates of their expected utility \citep{eikema-aziz-2020-map,eikema-aziz-2022-sampling}.
In this work, we will focus on this sampling-based approximation.\looseness-1%

\subsection{Structural Variation in Language Models}
The importance of modelling uncertainty in natural language generation systems has received growing attention in recent years \citep{baan2023uncertaintynaturallanguagegeneration}.
Crucially, uncertainty extends beyond surface-form variations in outcome space to encompass deeper variation in latent space.
To capture such variation, metrics like semantic entropy \citep{kuhn-etal-2023-semantic} and similarity-sensitive entropy \citep{cheng-vlachos-2024-measuring} have been proposed, primarily to identify when high uncertainty may signal potential model errors. 
Complementary work has examined similar measures with a different aim: to assess whether the uncertainty exhibited by language models aligns with the natural variability found in human-generated responses \citep{deng-etal-2022-model,giulianelli-etal-2023-comes,ilia2024variabilityneedimplyerror}. %

Recent applications of MBR have largely focused on neural machine translation---a relatively constrained task where, nonetheless, models have been shown to capture less variation than what human translators consider plausible \citep{giulianelli-etal-2023-comes}. 
Extending beyond translation, a few studies have applied MBR to other generation tasks. 
For example, \citet{suzgun-etal-2023-follow} successfully use BERTScore-based MBR for summarisation, data-to-text generation, textual style transfer, and image captioning. 
However, these tasks also tend to involve a limited range of plausible outputs. %
More recently, \citet{prometheus-mbr} applied MBR to instruction-following tasks, using an LLM-as-a-judge as a utility function. 
While this method yields strong results, it relies on a distillation step to approximate the utility, as directly querying an LLM judge during decoding is computationally prohibitive.
In this work, we propose three lightweight adaptations to standard similarity-based utility functions, specifically designed for open-ended tasks characterised by high variability in latent structure.

\begin{table*}[t]
\centering
\fontsize{10}{11.4}\selectfont
\rule{\textwidth}{0.08em}\vspace{0.6em}
\colorbox{blue!3}{%
    \parbox{15.3cm}{%
        \textbf{Prompt:} You are a support agent. The customer wrote: 
        ``My new Wi-Fi router keeps dropping the connection every hour. I have deadlines today.'' 
        Please respond in exactly ONE of these ways: 
        1) STEPS: give a concrete step-by-step fix; or 
        2) EMAIL: draft a warranty-replacement request email. Pick one format only.
    }
}

\vspace{0.6em}

\rowcolors{2}{gray!10}{white} %
\begin{tabular}{p{2cm} p{10.8cm} >{\centering\arraybackslash}p{1.9cm}}
\toprule
\textbf{Type} & \textbf{Candidate} & \textbf{Selected by} \\
\midrule
Instructions & Follow these steps: 1) Unplug the modem for 60s \ldots\ 2) Update firmware \ldots\ 3) Split SSIDs \ldots\ 4) Disable band steering \ldots\ 5) \ldots\ factory reset \ldots & -- \\
Instructions & Try this sequence: 1) Move the router \ldots\ 2) Install latest firmware \ldots\ 3) Turn off Smart Connect \ldots\ 4) Lock 5 GHz \ldots\ 5) Reboot and test \ldots & -- \\
Email & Subject: Router Disconnects — [Router Model] Hello [Store], My [Router Model], purchased on [date], disconnects every hour \ldots\ Please process a replacement \ldots\ Order \#[number], Serial [serial] \ldots & -- \\
Email & Subject: Warranty Replacement Request — [Router Model] Dear [Retailer], Since purchase on [date], this router drops Wi-Fi hourly \ldots\ I have attempted resets, firmware updates \ldots\ I am requesting a warranty replacement \ldots\ Order \#[number] \ldots & -- \\
Email & Subject: Faulty Unit — Replacement Needed Dear [Retailer], This router consistently drops Wi-Fi \ldots\ even after firmware updates, resets \ldots\ Kindly issue a replacement under warranty \ldots\ Order \#[number]; Serial [serial] \ldots & Structure-conditional MBR \\
Compromise & Here’s what you can try: 1) reboot the router, 2) check for firmware updates, and 3) split the Wi-Fi bands. Also, here's a line you could send to the store: ``My router disconnects every hour, please replace it.'' Hope one of these helps! & Standard MBR \\
\bottomrule
\end{tabular}
\caption{An example comparing standard BLEURT MBR with our proposed structure-conditional variant (Cosine). We prompt a customer service model to respond in one of two ways: either a set of troubleshooting instructions or a warranty replacement email. For brevity, generations are shortened with ``\ldots''. Standard MBR selects a compromise generation that mixes both formats, which is not optimal for either cluster and is undesirable given the prompt. In contrast, structure-conditional MBR opts for the cluster-optimal generation within the dominant cluster (\ie, the generation that standard BLEURT MBR would pick if it would only observe samples from the email cluster).}
\label{tab:motivating-example-generations}
\vspace{-0.1em}
\end{table*}

\section{Structure-Conditional Optimality}

The central question addressed in this paper is how commonly employed utility functions for MBR decoding behave when complex structural variation is present. In \cref{fig:continuous-bimodal-mbr}, we illustrate the problem with a simplified example, %
highlighting how the choice of utility function can influence decision-making---particularly when the outcome space contains multiple distinct modes.
In this example, the outcome space is modelled as a bimodal Gaussian, and the decision problem is to select a single ``best'' outcome on the real line. 
If we use the negative squared error as our utility function,\footnote{
    Equivalently, one may frame this as minimising the risk under a squared error loss function. %
} the theoretical optimum corresponds to the mean of the bimodal distribution (the light blue line in \cref{fig:continuous-bimodal-mbr}). 
This solution may be undesirable
as the mean lies in a region of low probability mass and is unlikely to be sampled in practice.
If we apply a sampling-based approximation to the decision rule, as is common in language generation applications of MBR, the approximation selects an outcome near this theoretical optimum, which typically resides at the boundary of one of the clusters. %
Alternatively, if we adopt a different utility function---such as a radial basis function kernel---the theoretical optimum shifts to the mode of the largest cluster (\cref{fig:continuous-bimodal-mbr}, dark blue line). 
This outcome, being more representative of a high-probability region, may be more desirable than either the low-probability intermodal mean or an outcome near the edge of a cluster.\looseness-1

In probability distributions over natural language, multiple such ``modes'' may also be present, albeit more difficult to define and detect. 
For example, generations might cluster around various semantically distinct plausible answers to a question, different intended dialogue acts in a response, or varying discourse structures.
Depending on the utility function used, this can result in behaviours analogous to those shown in \cref{fig:continuous-bimodal-mbr}. 
Whether a certain behaviour is desirable depends on the modeller; for instance, a between-cluster solution may be appropriate if the model assigns probability mass to responses like \textit{The answer could be either [A] or [B]}, but in other cases, it could lead to suboptimal decisions.  We illustrate this more concretely in \cref{tab:motivating-example-generations}, where we show an example in which standard MBR chooses an arguably suboptimal compromise between two clusters of valid responses.
In this work, we investigate this phenomenon and propose simple adaptations to utility functions that encourage behaviour more similar to that of the RBF utility in the continuous example. %

\subsection{Evaluating Structural Sensitivity in MBR}
\label{subsec:optimality-metrics}
To quantify the extent to which the MBR solution with commonly used utility functions respects structural variability in outcome spaces over natural language, we introduce two complementary metrics.
These metrics evaluate whether MBR solutions align with, or differ from, solutions obtained when conditioning on latent structures.

\paragraph{Cluster Optimality.} 
This metric quantifies the proportion of cases, over a test set, in which the MBR solution under the distribution $P(Y|x)$ matches the MBR solution under the conditional distribution $P(Y|x, s)$, where $s$ denotes an annotated structure (\eg, a dialogue act) that we additionally condition on.
Formally, let\looseness-1
\begin{equation} 
    \hat{y}_i = \argmax_{h} \E_{P(Y|x_i)}[\utility(h, Y)]
\end{equation}
be the MBR solution for input $i$, and
\begin{equation}
    \hat{y}_i^{(s)} = \argmax_{h} \E_{P(Y|x_i, s)}[\utility(h, Y)]
\end{equation}
the MBR solution conditioned on $s$. The cluster optimality metric is then defined, for test set $\dataset$, as
\begin{equation}
\text{CO} = \frac{1}{|\dataset|} \sum_{i \in \dataset} \indicator{\hat{y}_i = \hat{y}_i^{(s)}}
\label{eq:CO}
\end{equation}
where $\indicator{\cdot}$ is the indicator function.

\paragraph{Cluster-Optimal Rank Correlation.} In addition to the top-ranked solution, we also examine the full rankings produced by MBR. 
For each input $i$, consider a fixed set of hypothesis generations $\mathcal{H}_i^{(s)} = \{h_{i1}, \ldots, h_{in}\}$ corresponding to structure $s$. Define the rankings:
\begin{equation}
R_{ij} = \text{rank of } h_{ij} \text{ by } \E_{P(Y|x_i)}[\utility(h_{ij}, Y)] %
\end{equation}
\begin{equation}
R_{ij}^{(s)} = \text{rank of } h_{ij} \text{ by } \E_{P(Y|x_i,s)}[\utility(h_{ij}, Y)]
\end{equation} %
The cluster-optimal rank correlation is then the average Spearman's rank correlation coefficient $\rho$ between these two rankings over the test set:
\begin{equation}
\text{CORC} = \frac{1}{|\mathcal{D}|} \sum_{i \in \dataset} \rho\big( R_i, R_i^{(s)} \big)
\label{eq:CORC}
\end{equation}

\section{Standard Utility Functions are Not Structure-Conditionally Optimal}
\label{sec:utilities-are-not-optimal}

We now demonstrate that MBR solutions derived using standard utility functions, such as BERTScore or BLEURT, often diverge from those obtained when conditioning on latent structures. 
While this divergence may be acceptable from the perspective of the modeller, our analysis assumes %
a language production process in which the speaker first selects a latent structure---implicitly or explicitly---and then realises it through an utterance. 
Under this assumption, a generation should be optimal with respect to some latent structure, specifically the one selected by the speaker.\footnote{Note that we do not model the initial stage of this process, \ie, the selection or planning of the latent structure. 
Instead, we take it as given and focus on the requirement that the resulting generation be optimal within plausible realisations of the chosen structure.}

To investigate how sensitive the MBR solution is to structural uncertainty in the outcome space, we consider three representative types of latent structure---dialogue act, emotion, and response structure---each of which defines a plausible axis of variation in generated text (see \Cref{subsec:structures}).
For each structure, we construct a dataset that reflects the outcome space of a hypothetical model with uncertainty over that structure's possible instantiations.
We then compute the standard MBR solution over the entire outcome space, and assess its optimality using the evaluation criteria introduced in \cref{subsec:optimality-metrics}. %
The results of this analysis are summarised in \cref{tab:optimality-standard-mbr} and presented in \cref{subsec:optimality-results}.\looseness-1

\subsection{Constructing Outcome Spaces with Controlled Structural Uncertainty}
\label{subsec:structures}
We ground our analysis %
in three types of latent structure. 
This section defines each structure type and describes how we construct datasets to model uncertainty over their possible instantiations.

\subsubsection{Types of Latent Structure}
\label{subsec:latent-structure-types}
We examine three types of latent structure that are representative of structural variability in the outcome spaces of open-ended generation tasks.

\paragraph{Dialogue Act.} A dialogue act represents the communicative function or intent of an utterance within the context of a conversation. Following the taxonomy proposed by \citet{amanova-etal-2016-creating}, we focus on four dialogue act types: \textsc{inform}, \textsc{question}, \textsc{directive}, and \textsc{commissive}. %

\paragraph{Emotion.}
Another latent factor that shapes the form of an utterance in conversation is the emotion the speaker aims to express. In this work, we adopt Ekman's six basic emotions~\citep{ekman-emotions}: \textsc{happiness}, \textsc{sadness}, \textsc{fear}, \textsc{anger}, \textsc{surprise}, and \textsc{disgust}. These emotional states influence both lexical choice and broader stylistic features. %

\paragraph{Response Structure.}
This structure type captures how information is organised within an instruction-following response. We consider four ad-hoc categories: \textsc{brief}, a %
single-sentence reply; \textsc{paragraph}, a more developed, single-paragraph answer; \textsc{list}, a bullet-pointed set of items; and \textsc{table}, a structured tabular presentation.

\subsubsection{Dataset Construction}
\label{subsec:dataset-construction}

For each type of latent structure, we construct a dataset that simulates the outcome space of a hypothetical model with uncertainty over possible instantiations of that structure. 
We randomly sample conversational contexts from the DailyDialog corpus \citep{li2017dailydialog}---1,000 each for dialogue act and emotion---and take the first 1,000 instructions from the Alpaca dataset \citep{alpaca} for response structure.
We then prompt the instruction-tuned, 13B parameter variant\footnote{\texttt{allenai/OLMo-2-1124-13B-Instruct}} of the OLMo 2 model suite \citep{olmo2} to generate outputs for each category within each structure type, using hand-curated prompts (see \cref{sec:data-generation} for details). 
For every context, we generate 25 responses per structure category (\eg, 25 \textsc{brief}, 25 \textsc{paragraph}, 25 \textsc{list}, and 25 \textsc{table} responses).
This procedure results in 3,000 distinct outcome spaces, corresponding to 350,000 candidate generations in total.
In \cref{tab:cluster-optimal-examples} (\cref{app:dataset-examples}), we provide examples from the dataset, contrasting standard MBR solutions with cluster-optimal ones.

\begin{table}
\resizebox{\linewidth}{!}{%
    \centering
    \begin{tabular}{llcccc}
    \toprule
    \textbf{Metric} & \textbf{Utility} & \textbf{Dial.\ Act} & \textbf{Emotion} & \textbf{Resp.\ Str.} & \textbf{All (Avg)} \\
    \midrule
    \multirow{2}{*}{CO} 
        & BERTScore & 0.370 & 0.330 & 0.390 & 0.363 \\
        & BLEURT    & 0.410 & 0.510 & 0.530 & 0.483 \\
    \addlinespace
    \multirow{2}{*}{CORC} 
        & BERTScore & 0.081 & 0.084 & 0.080 & 0.082 \\
        & BLEURT    & 0.144 & 0.155 & 0.123 & 0.141 \\
    \bottomrule
    \end{tabular}%
}
    \caption{Cluster Optimality (CO) and Cluster-Optimal Rank Correlation (CORC) of MBR solutions obtained using BERTScore and BLEURT utility functions over constructed outcome spaces. %
    }
    \label{tab:optimality-standard-mbr}
\end{table}

\subsection{Structural Sensitivity of Standard MBR Utility Functions}
\label{subsec:optimality-results}

\Cref{tab:optimality-standard-mbr} presents cluster optimality (CO, Eq.~\ref{eq:CO}) and cluster-optimal rank correlation (CORC, Eq.~\ref{eq:CORC}) scores for MBR solutions under two standard utility functions across our three types of latent structure.
These metrics
quantify how often the MBR-selected response is optimal with respect to the latent structure (CO), and how well it aligns with the structure-optimal ranking (CORC).\looseness-1

Across all structure types, we observe a consistent degree of suboptimality. 
The CO scores indicate that in fewer than half of the cases, the MBR solution is optimal with respect to its underlying structure (36.3\% using BERTScore, 48.3\% with BLEURT).
This misalignment persists across dialogue act, emotion, and response structure, with no evident correlation to the number of clusters involved. This suggests that the failure to recover structure-optimal responses is not merely a consequence of increased structural granularity.
Moreover, while slight differences are present between BLEURT and BERTScore, both utility functions consistently select suboptimal generations and yield relatively weak ranking correlation. 
Overall, this analysis shows that standard utility functions possess low sensitivity to structural uncertainty.\looseness-1

\section{Structure-Conditional MBR Decoding}
\label{sec:structure-conditional-mbr}
To address the limitations of MBR decoding with standard utility functions in the presence of latent structural variability, we propose three structure-aware decoding approaches.

\paragraph{Utility Cut-off.}
Standard utility functions may implicitly penalise structural mismatches, but they do not prevent structurally dissimilar candidates from influencing the ranking of outputs.
To mitigate this, we introduce a simple utility cut-off mechanism that filters out low-utility comparisons when computing expected utility. 
Specifically, we modify the utility function $\utility(y, y')$ as follows:
\begin{equation}
\utilitycut(y, y') = 
\begin{cases}
\utility(y, y') & \text{if } \utility(y, y') \geq \threshold, \\
\cutoffconstant & \text{otherwise}
\end{cases}
\end{equation}
where $\threshold$ is a threshold fixed across the dataset, %
and $\cutoffconstant$ is a small constant (or zero).
This limits the influence of distant or structurally irrelevant samples,  %
aligning the MBR solution more closely with local modes in the outcome distribution.

\paragraph{Clustering.}
A more explicit approach to structure-aware decoding is to first partition the outcome space into clusters---each corresponding to a distinct latent structure---and then apply MBR within the dominant cluster. %
We implement this by clustering candidate generations using sequence embeddings $\embed(y)$ derived from a model~$\embed$ fine-tuned to detect particular structures of interest (\eg, dialogue act, response structure, or affective content). %
Formally, let $\mathcal{H} = \{h_1, \dots, h_n\}$ be the set of candidates, and let $\cluster_1, \dots, \cluster_k$ denote the resulting clusters, with $\candidates = \bigcup_{j=1}^k \cluster_j$.
At inference time, we restrict MBR decoding to the members of the largest cluster $\cluster^\star = \argmax_{\cluster_j} |\cluster_j|$ such that 
\begin{equation}
\hat{y}_{\textrm{cl}} = \argmax_{h \in C^\star} \; \E_{P(Y|x)}[\utility(h, Y) \mid Y \in \cluster^\star]
\end{equation}
To recover a full ranking over candidates (\eg, for evaluation), we first rank clusters by size, and then rank candidates within each cluster based on expected utility.
This two-stage approach prioritises high-utility responses as judged against structurally consistent pseudo-references, reducing the risk of inter-modal averaging in the selected outputs.\looseness-1

This procedure could theoretically also be formulated as an adaptation of the utility function: 
\begin{align}
\utility_\textrm{cl}(y, y') = &\, \indicator{\cluster(y) = \cluster(y')} \times \nonumber \\ 
& \times \utility(y, y') \times \indicator{\cluster(y) = \cluster^*}
\label{eq:cluster-mbr-u}
\end{align}
where $\cluster^*$ represents the cluster with highest probability mass under $P(Y|x)$.
Decoding then becomes standard MBR maximisation of expected utility under the adapted utility function.

\paragraph{Structure Embeddings.}
As an alternative to explicit clustering, we propose incorporating structural sensitivity into the utility function by leveraging structure-aware sequence embeddings. 
Specifically, we fine-tune a sequence embedding model~$\embed$ to encode the structural property of interest %
and redefine the utility function to weight candidate comparisons by candidate similarity in this embedding space.
Formally, for a candidate $y$ and a reference $y'$, we compute the modified utility as:
\begin{equation}
\utility_\textrm{emb}(y, y') = \utility(y, y') \cdot \cos\big( \embed(y), \embed(y') \big)
\end{equation}
where $\utility(y, y')$ is the original utility and $\cos(\cdot)$ denotes the cosine similarity between structure-sensitive embeddings.
To further reduce the influence of structurally mismatched samples, we also experiment with a threshold on cosine similarity: values below the threshold are set to zero, removing the contribution of the utility comparison to the expected utility altogether. 
In contrast to the Clustering approach, Structure Embeddings allow us to softly bias the MBR solution toward structurally coherent outputs without requiring the prediction of hard labels, potentially leading to greater robustness against imperfections in the clustering model.

\section{Experiments}
\begin{figure*}
    \centering
    \includegraphics[width=0.97\linewidth]{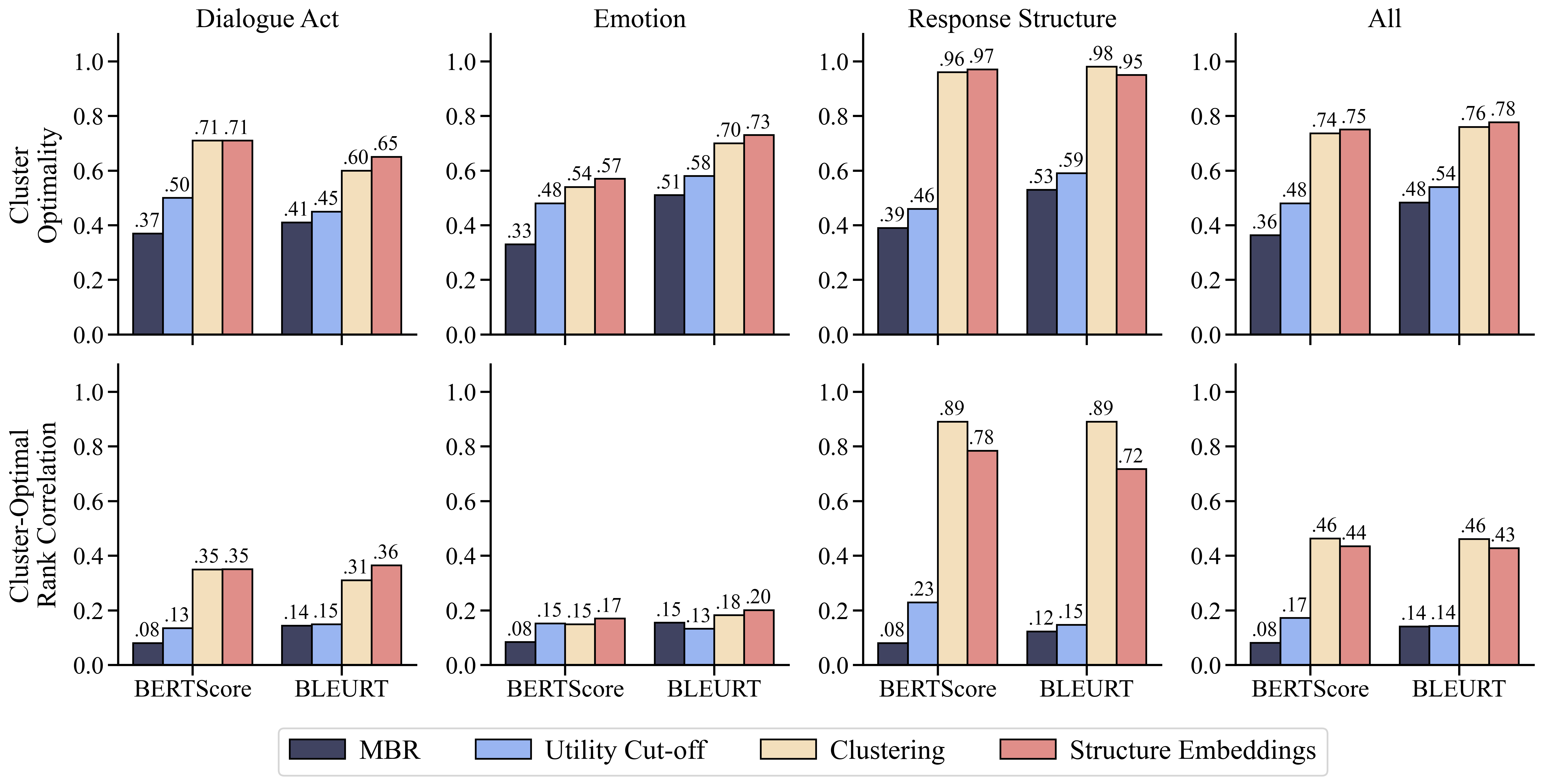}
    \caption{Cluster Optimality and Cluster-Optimal Rank Correlation on the constructed outcome spaces of \cref{subsec:structures}. We compare standard BERTScore and BLEURT MBR with the three adaptations to the utility functions proposed in \cref{sec:structure-conditional-mbr}.}
    \label{fig:controlled-structural-uncertainty}
\end{figure*}

To evaluate the effectiveness of the proposed methods, we conduct a series of experiments on the dataset we constructed in \cref{sec:utilities-are-not-optimal}, as well as two real-world instruction-following datasets. All our experiments use either BERTScore or BLEURT as the base utility function, two commonly employed utility functions in natural language generation \citep{freitag-etal-2022-high,suzgun-etal-2023-follow}.

\subsection{Cluster Optimality Under Controlled Structural Uncertainty}
We first assess our methods on the three datasets constructed in \cref{sec:utilities-are-not-optimal}, which contain generations consisting of various types of structural uncertainty: over dialogue acts, emotions, and response structures. 
Recall that we treat these generations as hypothetical outcome spaces of a language model. 
That is, we consider all generations for a given context to be unbiased samples from a language model that we wish to perform MBR decoding with. 
We split the 1,000 contexts in each dataset into training, validation, and test sets using an 800/100/100 split.\looseness-1

\paragraph{Hyperparameter Selection.} For each method proposed in \cref{sec:structure-conditional-mbr}, we use the training and validation splits to select hyperparameters and train the sequence embedding models. The threshold in the Utility Cut-off approach is optimised separately for BERTScore and BLEURT, resulting in different thresholds.
We base our sequence embedding models on the \texttt{all-mpnet-base-v2}\footnote{\url{https://huggingface.co/sentence-transformers/all-mpnet-base-v2}} Sentence Transformer  \citep{reimers-gurevych-2019-sentence}, which we further fine-tune using a triplet loss and gold annotations of underlying structure to enhance sensitivity to the structural variation present in our datasets. We use the same sequence embedding models for our Clustering and Structure Embeddings approaches. We find that jointly fine-tuning and selecting thresholds on the combination of all three types of latent structure leads to the most robust performance in terms of CO,\footnote{Generally, we find CO and CORC in validation procedures to align reasonably well.} and we use the resulting settings for the experiments below.
Further details on the hyperparameter selection and fine-tuning procedures can be found in \cref{app:hyperparameter-selection}.\looseness-1  %

\paragraph{Results.} We compare each of our proposed methods against standard sampling-based MBR decoding using either BERTScore or BLEURT as the utility function, and measure both cluster optimality (CO, Eq.~\ref{eq:CO}) and cluster optimal ranking correlation (CORC, Eq.~\ref{eq:CORC}). 
Results are shown in \cref{fig:controlled-structural-uncertainty}. 
All the methods we proposed improve cluster optimality compared to the baseline utility functions. %
Utility Cut-off yields the smallest improvement over standard BERTScore and BLEURT MBR, on average increasing CO by 11.7\% and 5.6\%, respectively, and CORC by 0.091 and 0.002, respectively. 
The Clustering and Structure Embeddings approaches perform considerably better than the baseline MBR. 
Clustering improves CO on average by 37.3\% / 27.7\% and CORC by 0.382 / 0.320 over standard BERTScore and BLEURT MBR, respectively. 
Similarly, Structure Embeddings improve CO on average by 38.7\% / 29.3\% and CORC by 0.354 / 0.287. %
We note that higher CO does not always correspond to higher CORC, indicating that achieving the cluster-optimal MBR solution is generally easier than recovering the entire ranking accurately.
Additionally, we observe that some types of latent structure are more difficult to capture effectively than others.

\subsection{Instruction-Following}
\label{sec:experiments-instruction-following}
    
\begin{table}
\resizebox{\linewidth}{!}{%
    \centering
    \begin{tabular}{lcccc}
    \toprule
    \textbf{Benchmark} & \textbf{MBR} & \textbf{Cut-off} & \textbf{Cluster} & \textbf{Embeddings} \\
    \midrule
    AlpacaEval         & 96.5\%       & 96.1\%          & $\mathbf{97.0\%}$           & 96.1\%          \\
    MT-Bench (single)  & 76.3\%       & $\mathbf{90.0\%}$          & 80.0\%           & 78.8\%          \\
    MT-Bench (multi)   & 71.3\%       & 70.0\%          & 72.5\%           & $\mathbf{74.4\%}$          \\
    \bottomrule
    \end{tabular}%
}
    \caption{AlpacaEval and MT-Bench Prometheus win rates versus \texttt{text-davinci-003} (AlpacaEval) / GPT-4o (MT-Bench). We compare standard BERTScore MBR with the approaches introduced in \cref{sec:structure-conditional-mbr}: Utility \textbf{Cut-off}, \textbf{Cluster}ing and Structure \textbf{Embeddings}.
    }
    \label{tab:prometheus}    
\end{table}
Next, we evaluate our methods on two real-world instruction-following datasets: AlpacaEval \citep{alpaca_eval} and MT-Bench \citep{zheng2023judgingllmasajudgemtbenchchatbot}.
In this case, we do not have access to any labelling of potential latent structure.  
We use the same hyperparameters and sequence embedding models from the previous set of experiments, tuned on the combination of all three datasets from \cref{sec:utilities-are-not-optimal}. 
As a language model, we select OLMo 2 (13B) \citep{olmo2}, and obtain 30 unbiased samples per prompt for use in MBR decoding.
To measure task performance, we use Prometheus\footnote{\texttt{prometheus-eval/prometheus-7b-v2.0}} \citep{kim2024prometheus} as a judge, conducting relative grading against \texttt{text-davinci-003} and GPT-4o \citep{openai2024gpt4ocard}, for AlpacaEval and MT-Bench, respectively.\footnote{   
    We did not find any available multi-turn system generations for the full MT-Bench dataset. Therefore, we generated our own from OpenAI's GPT-4o, using greedy decoding. 
} 
All experiments employ BERTScore as the base utility.
Further details on the generation and evaluation procedures are provided in App.~\ref{app:instruction-following}.

\paragraph{Results.}
\Cref{tab:prometheus} reports win rates against \texttt{text-davinci-003} and GPT-4o for standard MBR decoding with a BERTScore utility, alongside our structure-conditional utilities from \cref{sec:structure-conditional-mbr}. %
On AlpacaEval, the Clustering method outperforms standard MBR. In the single-turn MT-Bench setting, both Clustering and Utility Cut-off surpass standard MBR, with Utility Cut-off achieving a notable 13.7 percentage point improvement and reaching a 90\% win rate over GPT-4o. This indicates responses are often judged clearer, more helpful, accurate, and fully aligned with the intended purpose of the instruction.
Performance declines across the board in the more challenging multi-turn MT-Bench setting. 
However, both Clustering and Structure Embeddings continue to outperform standard MBR, demonstrating improved structural sensitivity also in extended interactions. 
Smaller gains here may stem from reduced uncertainty as conversational context accumulates, resulting in less diverse outcome spaces.
In such cases, structure-conditional utilities likely yield results similar to standard MBR, reducing the relative benefit of structural adaptations.
We also observe that Structure Embeddings tend to outperform Clustering, possibly because soft partitioning better captures subtle structural differences, whereas hard clustering might inadvertently exclude partially similar candidates.
Nevertheless, the lower overall MBR performance in multi-turn tasks suggests that these scenarios are inherently more challenging, beyond the effect of reduced variability.\footnote{To investigate whether structure-conditional MBR offers greater benefits in high structural-variability cases, we attempted to bin test items based on their structural variability (measured as average dispersion of structure embeddings from the fine-tuned embedding model); however, this analysis did not reveal any clear trends.} %

Overall, the consistent improvements of structure-aware MBR methods over standard MBR suggest that incorporating latent structural information not only enhances the theoretical optimality of MBR solutions but also improves generation quality in practical settings.

\section{Conclusion}
In this work, we examined the limitations of MBR decoding in open-ended generation scenarios, where outcome spaces might exhibit high structural variation.  %
We hypothesised that commonly used utility functions are insufficiently sensitive to latent structural uncertainty, leading to suboptimal generation choices within structurally coherent clusters of responses. %
To test this hypothesis, we constructed a dataset featuring naturally occurring contexts paired with outcome spaces that exhibit controlled variation in dialogue act, emotion, and response structure. 
Our findings confirm that MBR decoding under standard utilities frequently fails to select cluster-optimal candidates, with suboptimal selections occurring in more than half of the cases.%

To address this issue, we proposed three approaches to adapt utility functions to be more structurally aware. 
The corresponding approaches---Utility Cut-off, Clustering, and Structure Embeddings---demonstrate significant improvements in both cluster optimality (reaching up to 98\% for response structure) and cluster-optimal rank correlation (reaching up to 0.89 for response structure). %
Importantly, these methods incur only modest additional computational cost, requiring only lightweight fine-tuning of a sequence embedding model or performing a hyperparameter search for a threshold value.
Once optimised, they can be applied directly to unlabelled data. Our experiments indeed show improvements in generation quality on the real-world benchmarks AlpacaEval and MT-Bench without additional supervision. 
While further investigation is required to clarify the scope of these zero-shot capabilities, we speculate that structure embeddings may generalise because they capture structures similar to those seen during fine-tuning (\eg, other emotion categories) or leverage features the model was already sensitive to from pre-training (\eg, semantically varying generations). Joint fine-tuning on multiple structure types may further enhance this type of generalisation, potentially enabling the model to handle entirely unseen structural variations.

Based on our positive results in both controlled and real-world settings, we recommend adopting structure-aware MBR decoding in tasks characterised by medium to high outcome space variability, such as instruction-following and conversational tasks. 
We encourage future research into structure-sensitive utility functions that build on this work to achieve even greater cluster optimality, generation quality, or inference-time efficiency. %
We also see value in further investigating the relationship between outcome space variability and the effectiveness of structure-aware MBR, as well as between cluster optimality and overall generation quality.\looseness-1 %

\section*{Limitations}
To test our hypothesis on the suboptimality of standard similarity-based MBR utility functions, we relied on a curated dataset that captures three representative types of latent structure commonly found in open-ended natural language generation tasks. However, this dataset does not exhaustively cover all possible structural variations present in natural language. 
Additionally, our evaluation assumes that language models accurately represent uncertainty over latent structures---an assumption that may not always hold in practice \citep[see, e.g.,][]{giulianelli-etal-2023-comes}. For example, in a dialogue setting, a model might assign most of its probability mass to responses aligned with the \textsc{inform} dialogue act category, even though human responses would display a broader range of structural types. 
As discussed in \cref{sec:experiments-instruction-following}, we tried binning test items according to their measured structural variability to assess whether structure-conditional MBR provides greater benefits in high-variability cases. This analysis, however, did not yield any clear trends.
Furthermore, while our study focuses on standard similarity-based utility functions such as BLEURT and BERTScore, we acknowledge that task-specific or learned reward models could serve as alternative MBR utilities. Exploring how such utilities behave in the presence of structural variation is a promising direction for future work.

In terms of computational requirements, our methods introduce minimal overhead beyond standard MBR decoding. It is worth noting, however, that MBR decoding itself \textit{is} significantly more computationally demanding than greedy decoding or sampling a single generation. 
Since our approaches build on MBR, they inherit this higher computational cost. 
Nevertheless, they stand to benefit from recent advances aimed at improving the efficiency of MBR decoding \citep{cheng-vlachos-2023-faster, vamvas-sennrich-2024-linear, yang-etal-2024-direct}.

Finally, in our evaluation on instruction-following datasets, we rely on Prometheus as an LLM judge  \citep{kim2024prometheus}. LLM judges are imperfect evaluators, may be biased towards particular types of responses \cite{wang-etal-2024-large-language-models-fair,stureborg2024inconsistent}---for example, longer or more elaborate ones---and do not always align with human judgements \cite{zeng2024evaluating,bavaresco2024judges}. Additionally, Prometheus relies on a predefined rubric, and its performance may be sensitive to the specific formulation of that rubric. We did not conduct extensive experiments with alternative rubric designs, which may influence the robustness of the results.

\section*{Acknowledgments}
\begin{wrapfigure}[2]{l}{0.12\linewidth}
\vspace{-15pt}
\includegraphics[width=0.08\textwidth]{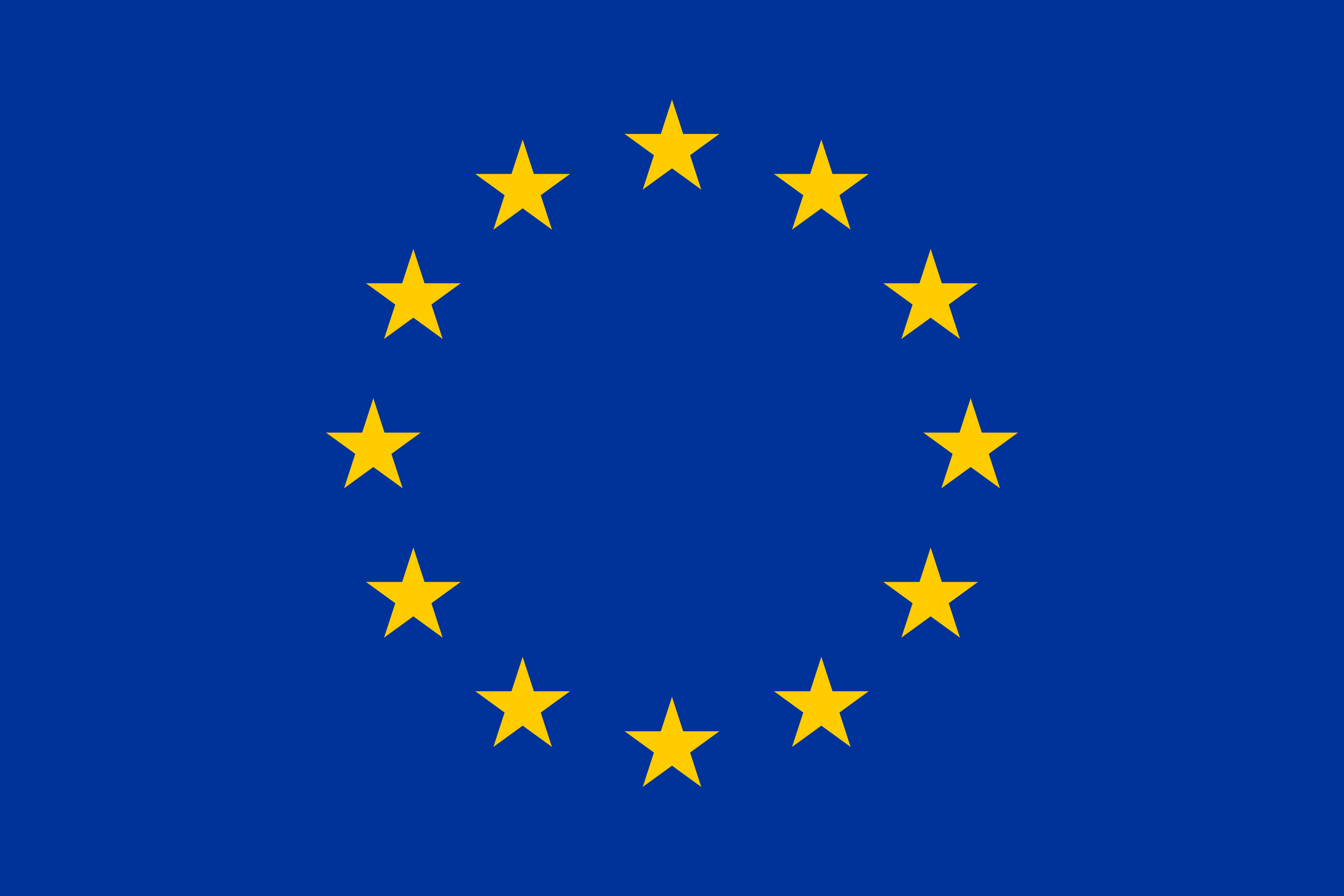}
\end{wrapfigure} 
This project received funding from the European Union’s Horizon Europe Research and Innovation programme under Grant Agreement No 101070631 (UTTER). Mario Giulianelli was supported in part by an ETH Zurich Postdoctoral Fellowship. We thank Ryan Cotterell for his valuable suggestions during the early stages of this project, and the anonymous ARR reviewers for their helpful feedback. 

\bibliography{anthology,custom}
\bibliographystyle{acl_natbib}

\appendix
\section{Data Generation}
\label{sec:data-generation}
In \cref{subsec:dataset-construction}, we semi-automatically construct a natural language dataset of hypothetical outcome spaces with varying underlying latent structure: dialogue act, emotion, and response structure. Here, we describe the full generation procedure in more detail.

\subsection{Generation Procedure}

We use dialogue contexts from the DailyDialog dataset \cite{li2017dailydialog} as a basis for the dialogue act and emotion subsets,
and prompts from the Alpaca \citep{alpaca} dataset as a basis for the response structure subset. We sample generations from the 13B OLMo 2 model \citep{olmo2}.\footnote{\texttt{allenai/OLMo-2-1124-13B-Instruct}}

\begin{enumerate}
    \item \textbf{Pre-processing.} 
    We preprocess DailyDialog by removing extra white spaces and keeping only dialogues with at least two turns. For each dialogue, we then randomly select a number of turns to include, chosen between two and the total number of turns minus two. Each turn is prefixed with \textit{"A:"} or \textit{"B:"} to indicate the speaker. For the Alpaca dataset, we simply discard prompts that contain an additional \texttt{input} field.
    \item \textbf{Prompt creation}. We randomly select 2,000 dialogues from the DailyDialog dataset---1,000 for dialogue act and 1,000 for emotion chosen independently---and the first 1,000 instructions without \texttt{input} field from the Alpaca dataset. We then fill out the predefined prompt templates (as defined in \cref{subsec:prompt_templ}) with the selected examples, resulting in prompts for each category within every latent structure. In total, we create 1,000 $\times$ 4 $=$ 4,000 inputs for the dialogue act subset, 1,000 $\times$ 6 $=$ 6,000 inputs for the emotion subset, and 1,000 $\times$ 4 $=$ 4,000 inputs for response structure subset.
    \item \textbf{Generation}. Using OLMo 2 (13B), we then generate 25 unbiased samples for each of the 14,000 constructed input prompts.
    \item \textbf{Post-processing}. We finalise the procedure by removing quotation marks around the generation and stripping any \textit{"A:"} or \textit{"B:"} prefixes at the start of generations.
\end{enumerate}

\subsection{Prompt Templates} \label{subsec:prompt_templ}

To generate prompts covering all categories within a latent structure, we define three types of prompt templates---one for each latent structure.

\subsubsection{Dialogue Act} \label{subsec:prompt_templ_DA}

For the dialogue act subset, we define the \texttt{responder} as the speaker whose turn it is now to speak (\eg, if the dialogue excerpt ends at \texttt{A}'s turn, we define \texttt{responder} $=$ \texttt{B}). For each \texttt{dialogue from DailyDialog}, we then iterate through the four defined dialogue acts and pass each of them as \texttt{act\_name}.

\begin{lstlisting}

### **Types of Dialogue Acts**
Here are common categories of dialogue acts, though exact categorizations may vary depending on the framework:
#### **1. Inform**
   - The Inform class contains all statements and questions by which the speaker shares information with the listener. The speaker assumes the information is correct and believes the addressee does not know or is not aware of it yet.
   - **Examples**:  
     - "The meeting starts at 3 PM."  
     - "I've already emailed the report."
     - "I saw John at the store yesterday."
     - "William Shakespeare wrote it."
     - "The Eiffel Tower is in Paris."

#### **2. Question**
   - The Question class includes speech acts where the speaker seeks information by asking a question. These acts are used when the speaker wants to know something and believes the listener has the answer. Questions can take different forms, including Propositional Questions (yes/no questions), Check Questions (confirming known information), Set Questions (open-ended questions), and Choice Questions (questions with multiple options).
   - **Examples**:  
     - "Did you finish your assignment?"
     - "You've met Sarah before, haven't you?"
     - "What time does the meeting start?"  
     - "Could you clarify what you meant by that?"
     - "Do you prefer coffee or tea?"

#### **3. Directive**
   - The Directive class includes speech acts where the speaker wants the listener to perform an action. This class covers Requests (asking someone to do something), Instructions (giving direct orders or guidance), Suggestions (offering recommendations), and Accepting or Rejecting Offers (responding to proposals). These acts differ based on how much pressure the speaker applies and their assumptions about the listener's willingness and ability to comply.
   - **Examples**:  
     - "Can you send me the file?"  
     - "Fill out this form before the appointment."
     - "You should try the new Italian restaurant downtown."
     - "Yes, I'd love to join you for dinner!"
     - "No, I can't take on another project right now."

#### **4. Commissive**
   - The Commissive class involves speech acts where the speaker commits to performing an action in the future. These acts include Accepting or Rejecting Requests, Suggestions, and Offers. By performing a Commissive act, the speaker is making a promise or commitment to carry out the action requested, suggested, or offered. These acts reflect the speaker's willingness to take responsibility for fulfilling the commitment, whether by agreeing to a proposal or refusing it.
   - **Examples**:  
     - "Fine, I'll pick you up at 5 PM."
     - "Sorry, I can't do that right now."
     - "That sounds great, I'll take the promotion."
     - "I promise to finish the report by the end of the day."
     - "I'll make sure to take care of it this weekend."

---

### **Dialogue Excerpt**

{dialogue from DailyDialog}

---

### **Instructions**
Please consider the provided dialogue excerpt and provide a plausible response (and only a single response) for {responder} that reflects the following dialogue act: {act_name}. Output only {responder}'s response with no additional text.<end_of_prompt>
\end{lstlisting}

\subsubsection{Emotion} \label{subsec:prompt_templ_E}

For the emotion subset, we again define the \texttt{responder} in the same way as in \cref{subsec:prompt_templ_DA}. For each \texttt{dialogue from DailyDialog} we then iterate through the six defined emotions and pass each of them as \texttt{emotion\_name}.

\begin{lstlisting}
    ### **Types of Emotions**
Here are seven main categories of emotions.

#### **1. Anger**
   - The Anger category represents emotions related to feelings of displeasure, hostility, or frustration. This emotion often arises when someone feels wronged or blocked from achieving their goal. It can range from mild irritation to intense rage.
   - **Examples**:  
     - "I can't believe this is happening!"  
     - "This is so unfair!"
     - "Why does everything always go wrong for me?"
     - "I'm so frustrated with this situation!"
     - "I'm really mad about how things turned out."

#### **2. Disgust**
   - The Disgust category includes emotions related to a strong sense of revulsion, disapproval, or distaste. It often arises when something is perceived as offensive, repellent, or morally objectionable.
   - **Examples**:  
     - "That food looks awful!"  
     - "I can't stand how they treat people."
     - "This is disgusting. I can't believe they did that."
     - "I feel sick just thinking about it."
     - "That's absolutely revolting!"

#### **3. Fear**
   - The Fear category includes emotions related to anxiety, nervousness, and concern about possible danger or harm. Fear can be rational or irrational and may cause physical or psychological distress.
   - **Examples**:  
     - "I'm really scared about what's going to happen."  
     - "I don't know if I can handle this situation."
     - "What if things don't go as planned?"
     - "I'm afraid something bad might happen."
     - "I'm nervous about the meeting this morning."

#### **4. Happiness**
   - The Happiness category includes emotions related to joy, contentment, and pleasure. Happiness is often associated with positive experiences, accomplishments, and satisfying events.
   - **Examples**:  
     - "I'm so excited about this weekend!"  
     - "This is such a great day!"
     - "I feel so happy about my progress."
     - "That sounds amazing, I'm really looking forward to it!"
     - "I'm so glad everything worked out!"

#### **5. Sadness**
   - The Sadness category represents emotions related to feelings of loss, disappointment, or sorrow. It often arises when there is a sense of unmet expectations, failure, or grief.
   - **Examples**:  
     - "I feel so down about what happened."  
     - "I can't stop thinking about it, it's just so upsetting."
     - "I'm really sad things turned out this way."
     - "It's been a tough time, and I feel heartbroken."
     - "I don't know how to get over this sadness."

#### **6. Surprise**
   - The Surprise category represents emotions related to unexpected events or outcomes, ranging from shock to awe. This emotion can be positive or negative, depending on the nature of the surprise.
   - **Examples**:  
     - "Wow, I didn't see that coming!"  
     - "That's such a surprise, I can't believe it!"
     - "I'm totally shocked by what happened."
     - "I wasn't expecting that at all!"
     - "I'm so surprised you did that!"

---

### **Dialogue Excerpt**

{dialogue from DailyDialog}

---

### **Instructions**
Please consider the provided dialogue excerpt and provide a plausible response (and only a single response) for {responder} that reflects the following emotion: {emotion_name}. Output only {responder}'s response with no additional text.<end_of_prompt>
\end{lstlisting}

\subsubsection{Response Structure} \label{subsec:prompt_templ_RS}
For the response structure subset, we define four different prompt templates, one for each category of response structure. For each \texttt{prompt from Alpaca}, we then append each of these templates, resulting in four different prompts---one per category---per input instruction.\\\newline
\textsc{Brief} 
\begin{lstlisting}
    {prompt from Alpaca} Give me a brief sentence with the answer. Make sure to restrict your response to a single sentence.
\end{lstlisting}
\textsc{Paragraph}
\begin{lstlisting}
    {prompt from Alpaca} Write an extensive paragraph on the topic. Restrict your answer to a single paragraph
\end{lstlisting}
\textsc{List}
\begin{lstlisting}
    {prompt from Alpaca} In your answer, make sure to include a bullet point list of items relevant to the topic. Keep your answer brief and make sure it contains a bullet point list.
\end{lstlisting}
\textsc{Table}
\begin{lstlisting}
    {prompt from Alpaca} In your answer, include a table relevant to the topic. Keep your answer brief and make sure it contains a table.
\end{lstlisting}

\section{Hyperparameter Selection}
\label{app:hyperparameter-selection}
We randomly split our generated datasets (dialogue act, emotion, and response structure) into 800/100/100 training/validation/testing data points. All data points consist of an input context and 25 generations per type of latent structure we are considering for that input (\eg, 25 generations each for \textsc{Brief}, \textsc{Paragraph}, \textsc{List}, and \textsc{Table} for a total of 100 generations). We compute BERTScore and BLEURT MBR solutions conditioned on each labelled cluster to get cluster-optimal rankings and MBR solutions to compare to. We use the training and validation splits for fine-tuning sequence embedding models and for hyperparameter selection. We perform all training and hyperparameter selection both on individual datasets (either dialogue act, emotion, or response structure) and on the combination of all datasets. We find that models trained on all data perform best overall and thus use these in our experiments. We proceed here to discuss the results of hyperparameter selection for each individual approach in more detail.

\paragraph{Utility Cut-off.} We considered both an absolute threshold on the utility value as well as a threshold on the deviation from the highest observed utility in the sample. We do not consider any utility comparisons with the candidate itself, \ie, we mask out the diagonal of the utility matrix. Furthermore, we experiment with setting utility values below the threshold to $0$ or $-1$, as well as discarding those utility comparisons altogether. We test a range of 50 threshold values ranging within reasonable values for the utility function itself, and order settings based on cluster optimality on the training data. We then take the 10 best-performing setups and select the one with the highest cluster optimality on the validation data. We tune the threshold independently for both BLEURT and BERTScore. We find an absolute value threshold to work best for both utilities, with values below the threshold zeroed out. We find an optimal threshold of 0.512 and 0.918 for BLEURT and BERTScore, respectively.

\paragraph{Clustering.} We use the Sentence Transformers \texttt{all-mpnet-base-v2} model as a basis for obtaining sequence embeddings. We further fine-tune this model using a triplet loss on triplets from our labelled datasets. We experiment with learning rates between $1 \times 10^{-4}$, $1 \times 10^{-5}$ and $1 \times 10^{-6}$, and find a learning rate of $1 \times 10^{-5}$ to lead to best validation loss overall. We then use these sequence embeddings with the $k$-means algorithm to obtain clusters. We select a number of clusters based on the silhouette score for $k=[2, 6]$ and set a threshold that the silhouette scores need to reach, otherwise $k$ is set to 1 and we consider all generations to come from a single cluster. This threshold is tuned based on prediction accuracy on the number of clusters for a range of values in $(0, 1)$, using random subsamples of the validation data with a random number  of clusters per subsample. 

\paragraph{Structure Embeddings.} Here, we use the same fine-tuned Sentence Transformer model from the Clustering approach. We shift and compress cosine similarity values to range between 0 and 1. We optionally consider a threshold on cosine similarity and perform an identical selection procedure to that for the threshold in the Utility Cut-off approach. We find that a threshold does considerably improve cluster optimality, with the best results obtained at a threshold of 0.918.

\paragraph{Fine-Tuned Utilities: BERTScore and BLEURT.} We also attempted fine-tuning BERTScore and BLEURT directly to be more sensitive to the latent structures we expect in the data. We experimented with fine-tuning BERTScore with a triplet loss on the sequence embeddings of the underlying \texttt{roberta-large} model, and used a mean squared error regression loss to fine-tune BLEURT to predict comparisons with out-of-cluster generations as 0 or -1. We attempted a range of hyperparameter values, but found that the resulting utility functions performed poorly across the board. Hence, we have not included those models in the main paper.

\section{Evaluation on AlpacaEval and MT-Bench}
\label{app:instruction-following}

\begin{figure}[t]
    \centering
    \begin{minipage}{0.9\linewidth}
    \begin{lstlisting}
You are an advanced AI assistant specializing in clear, well-reasoned, and articulate responses. Your goal is to provide comprehensive and accurate answers while ensuring coherence, logical consistency, and factual correctness. Be precise, provide evidence-based explanations, and use structured reasoning when appropriate. If a question has multiple interpretations, clarify them before answering. Avoid unnecessary verbosity while maintaining completeness. If uncertain, state your level of confidence and explain why.
    \end{lstlisting}
    \end{minipage}
    \caption{System prompt used, for all decoding methods, when generating for AlpacaEval and MT-Bench.}
    \label{fig:system-prompt}
\end{figure}

\noindent We conducted our evaluation of instruction-following generations on AlpacaEval and MT-Bench using Prometheus as an LLM-as-a-judge model\footnote{\texttt{prometheus-eval/prometheus-7b-v2.0}} and following this procedure:
\begin{enumerate}
    \item \textbf{Generation}. For each instruction from the dataset, we generate our answers from each respective decoding method.
    We use the system prompt given in \cref{fig:system-prompt}. When generating for single-turn MT-Bench, we only prompt the model with the first turn and store its output for evaluation. When generating for multi-turn MT-Bench, we first prompt the model with just the first turn, store its output for evaluation, and then we prompt it again with both turns and the reference GPT-4o generation to the first prompt.\footnote{We opted to always provide the reference response in multi-turn MT-Bench to avoid compounding errors.} The total number of instructions for multi-turn MT-Bench is thus twice the number of instructions for single-turn MT-Bench.
    \item \textbf{Evaluation}. We then pass the instruction, the reference answer, as well as the generations of our decoding methods to Prometheus.
    The reference answers for AlpacaEval (included in the dataset) were generated by \texttt{text-davinci-003}. For MT-Bench, we collected reference answers via greedy decoding from GPT-4o through the OpenAI API. We use the predefined \texttt{RELATIVE\_PROMPT\_WO\_REF} prompt template for Prometheus to obtain pairwise judgements---for each pair of competing outputs, Prometheus returns one letter (\textit{A} or \textit{B}) defining which output is preferred. We define the grading rubric as follows: \textit{``Is the answer clear, helpful, accurate, and fully aligned with the intended purpose of the instruction?''}
    \item \textbf{Final Score}. For every decoding method, we then calculate that method's win rate against the set of reference generations according to Prometheus. In multi-turn MT-Bench, we report the average of the win rates of both turns.
\end{enumerate}

\section{Dataset Examples}
\label{app:dataset-examples}
\cref{tab:cluster-optimal-examples} provides illustrative examples from the dataset described in \cref{subsec:structures}. It contrasts the solutions obtained with standard BERTScore-based MBR against the cluster-optimal BERTScore-based MBR solution. 

\begin{sidewaystable*}
\centering
\tablefoursize
\begin{tabular}{p{7cm} p{6.8cm} p{7cm} p{1.5cm}}
\toprule
\textbf{Context} & \textbf{Standard MBR Solution (BERTScore)} & \textbf{Cluster-Optimal MBR Solution (BERTScore)} & \textbf{Cluster label} \\
\midrule
A: Are you free this evening, Li?\newline
B: I’m afraid I’m busy tonight. I’m having dinner with friends.
& Sure, we can reschedule for another time then.
& Sure, maybe another time then. Let's catch up soon.
& Commissive \\
\midrule
A: Hi, Mike. We're having a party tonight, wanna come join us? You can bring your girlfriend.\newline 
B: Well. I am breaking up with Cathy.\newline 
A: What happened? Did you have a fight?\newline 
B: No. She is really a very nice girl...\newline 
A: Yeah, she is. She is pretty, caring, getting easy-going and she is a great cook. She is a gem.\newline 
B: Well, you are right. But sometimes she is too caring. Well, no... she’s just possessive and I kind of want a break... you know... for some room for myself.\newline 
A: Oh?! Did you talk with her about it?\newline 
B: I've tried, but it didn't work.\newline 
A: So, when are you going to tell her?\newline 
B: Maybe tonight. I don't know. I don't know how to tell her.\newline 
& Oh, Mike... I'm really sorry to hear that. This sounds really difficult for both of you. Just remember, it's important to communicate openly and honestly. 
& Oh, Mike... I'm really sorry to hear that. It must be tough dealing with this situation. 
& Sadness \\
\midrule
A: Well, I can’t believe your uncle and aunt came to stay with us for a week, and they did not even call us in advance.\newline 
B: John, they are old-fashioned people from the Chinese countryside. That’s their custom.\newline 
A: I know, but it is hard for me to get used to your relatives coming to visit us, and without warning.
& I'm actually really frustrated by this. Their lack of communication just throws everything off, and it's not fair to us! 
& How could they just show up like that? It's totally disrespectful and I'm furious!
& Anger \\
\midrule
Find the prime factorization for the number 22. 
& The prime factorization of 22 is 2 * 11, indicating it is not a prime number.
& The prime factorization of 22 is 2 * 11. 
&Brief \\
\midrule
Explain the theory of gravity.
& The theory of gravity, as described by Newton and later refined by Einstein's General Relativity, explains the attraction between objects with mass.
\begin{itemize}
    \item According to Newton, gravity is a force that attracts two bodies towards each other, proportional to their masses and inversely proportional to the square of the distance between them.
    \item Einstein's theory views gravity not as a force, but as a curvature of spacetime caused by mass. Massive objects cause a distortion in the fabric of spacetime, which guides the motion of other objects.
\end{itemize}
These theories collectively help us understand the predictable behavior of planets, stars, and other cosmic phenomena.
& The theory of gravity, primarily explained by Isaac Newton and later refined by Albert Einstein, describes the force that attracts two bodies with mass towards each other. Here are key points relevant to the theory of gravity:
\begin{itemize}
    \item Newton's Law of Universal Gravitation: \ldots %
    \item Influence of Mass and Distance: \ldots
    \item Orbital Mechanics: \ldots
    \item Equivalence Principle: \ldots
\end{itemize}
These concepts collectively provide a theoretical framework for understanding gravitational phenomena.
& List \\
\bottomrule
\end{tabular}
\caption{Examples from the dataset introduced in \cref{subsec:structures}. 
The table contrasts the solution selected by standard MBR (BERTScore-based utility) with the 
cluster-optimal MBR solution (BERTScore-based) for the given underlying structure. 
The last example was abbreviated with ``\ldots'' to fit within the table.}
\label{tab:cluster-optimal-examples}
\end{sidewaystable*}

\end{document}